\documentclass[11pt]{article}

\usepackage[preprint]{acl}

\usepackage{times}
\usepackage{latexsym}
\usepackage{amsmath}
\usepackage{algorithm}
\usepackage{algpseudocode}
\usepackage{amssymb}
\usepackage[T1]{fontenc}

\usepackage[utf8]{inputenc}

\usepackage{microtype}
\usepackage{booktabs}
\usepackage{inconsolata}
\usepackage{graphicx}
\usepackage[dvipsnames,table]{xcolor}
\usepackage[most]{tcolorbox}
\usepackage{tabularx}
\usepackage{soul}
\usepackage{enumitem}
\usepackage{tcolorbox}
\usepackage{xspace}
\tcbuselibrary{breakable}
\definecolor{HeaderGray}{gray}{0.92}
\definecolor{ColGray}{gray}{0.95}
\definecolor{RowGray}{gray}{0.97}
\usepackage{listings}
\definecolor{LightGreen}{HTML}{F0F5E9}
\setitemize[1]{itemsep=0pt,partopsep=0pt,parsep=0pt,topsep=0pt,leftmargin=12pt}
\newtcolorbox{problemBox}{
  colback=gray!10,
  colframe=gray!80,
  boxrule=0.5pt,
  arc=3pt,
boxsep=2pt,
left=0pt, right=0pt, top=0pt, bottom=0pt,
  width=\columnwidth,        
  before skip=10pt,
  after skip=10pt,
}

\newcommand{\model}{\textsc{PAIR-Med}\xspace}
\newcommand{\prm}{\textsc{PAIR-Prm}\xspace}

\title{\model: Policy–Reward Alternation for Iterative Refinement in Multilingual Medical Reasoning}

\author{
 \textbf{Fan Gao\textsuperscript{1}},
 \textbf{Xiaohang Xu\textsuperscript{1}},
 \textbf{Lang Huang\textsuperscript{2}},
 \textbf{Nan Lu\textsuperscript{3}},
 \textbf{Irene Li\textsuperscript{1}}
\\
\\
 \textsuperscript{1}The University of Tokyo,
 \textsuperscript{2}National Institute of Informatics,\\
 \textsuperscript{3}University of Bristol,
\\
 \small{
   Correspondence: \href{bkjl6178@g.ecc.u-tokyo.ac.jp}{bkjl6178@g.ecc.u-tokyo.ac.jp}
 }
}

\begin{document}
\maketitle
\begin{abstract}
Large language models demonstrate strong medical question-answering capabilities, yet reasoning performance remains uneven across languages. 
Process reward models (PRMs) provide step-level feedback for improving reasoning, but a fixed PRM may become increasingly misaligned with the trajectory distribution of an evolving policy during reinforcement learning, thereby reducing the reliability of its supervision.
We propose a policy--reward alternating training framework that jointly adapts a medical PRM and optimizes a multilingual reasoning policy. 
Following supervised initialization, the current policy generates multilingual trajectories to update the PRM, which then provides process-level rewards for the next policy update.
Medical step supervision maintains the PRM's medical foundation, while process, outcome, and language-related rewards guide policy learning. We develop multilingual medical PRMs (\prm) and reasoning models (\model) at 8B and 9B scales across 12 languages. 
On the medical subsets of Global-MMLU and MMLU-ProX, our models improve average accuracy by 3.01--7.47\% over their corresponding Qwen backbones. On CUREMED-BENCH, \model-9B improves the joint language consistency and logical accuracy metric by 12.88\% and 7.45\% over baselines for in-domain and out-of-domain languages. 
Further analyses examine PRM adaptation, multilingual reasoning quality, and the contributions of individual training components.
\footnote{The source codes are available at \url{https://anonymous.4open.science/r/PAIR-Med-475C/}}
\end{abstract}
\section{Introduction}

Medical reasoning requires integrating clinical observations with domain knowledge to derive well-supported conclusions.
Large language models (LLMs) have demonstrated strong performance on medical question answering, but expert evaluations continue to identify shortcomings in factuality, reasoning quality, and completeness~\cite{singhal2023large, singhal2025toward}.
A correct final answer does not necessarily imply valid reasoning; models may rely on unsupported inferences, overlook clinically relevant evidence, or produce internally inconsistent explanations~\cite{arora2025healthbench, griot2025large}.
Moreover, strong exam performance does not guarantee reliable guideline adherence or accurate interpretation of clinical evidence~\cite{hager2024evaluation}, while explanation-focused benchmarks further highlight the need to assess medical understanding beyond answer-level correctness~\cite{kim2024medexqa, yun2025med, jiang2026meds3}.
These findings motivate more direct evaluation and improvement of the medical reasoning process.
\begin{figure}
    \centering
    \includegraphics[width=1.1\linewidth]{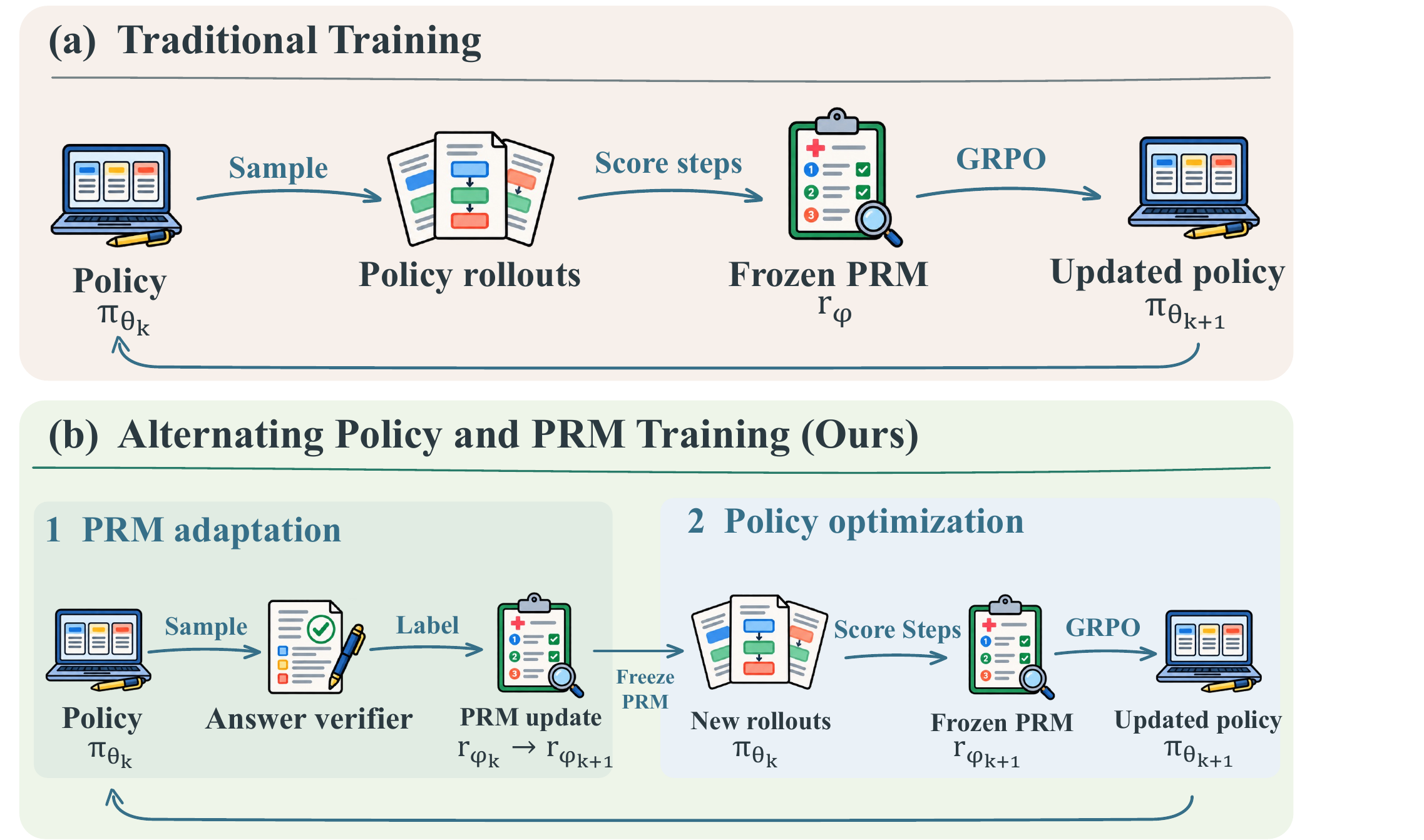}
    \caption{Traditional Frozen-PRM Training vs. Alternating Policy–PRM Training in \model.}
    \label{fig:comparison}
\end{figure}

Multilingual medical reasoning requires models to understand medical questions and produce answers in diverse languages.
Although multilingual medical models have expanded language coverage~\cite{qiu2024towards, zheng2025efficiently}, their reasoning performance remains uneven, particularly for languages with limited resources~\cite{gao-etal-2026-med, onyame-etal-2026-cure, chen2026healmed}.
Recent multilingual medical benchmarks have incorporated rationales, expert-written explanations, and long-form reasoning to characterize model behavior beyond answer accuracy~\cite{alonso2024medexpqa}.
These efforts highlight the importance of understanding how medical reasoning is expressed and preserved across languages.

Process reward models (PRMs) provide step-level feedback for evaluating and improving reasoning trajectories~\cite{lightman2024let, wang2024math}.
Recent work grounds medical process assessment in clinical evidence~\cite{yun2025med} and extends process supervision to multilingual reasoning using translated annotations~\cite{wang2025demystifying}.
However, during policy optimization, shifts in generated reasoning can expose errors in a fixed reward model and reduce the reliability of fixed process feedback~\cite{eisenstein2023helping, rafailov2024scaling, cui2025process}.
In multilingual medical reasoning, useful process supervision therefore requires assessing medical reasoning in each language and remaining responsive to the trajectories generated during training.
Building on these advances, we study 

\begin{center}
\begin{problemBox}
\begin{quote}
    \textit{How can medical process supervision adaptively guide accurate, consistent reasoning across languages?}
\end{quote}
\end{problemBox}
\end{center}

We introduce an alternating training framework in which a shared medical PRM and a multilingual reasoning policy learn from each other across successive rounds.
We first initialize both models with supervised multilingual medical reasoning data.
In each round, the policy generates multilingual trajectories to update the PRM, and the updated PRM provides process feedback for the next policy update.
Multilingual medical step supervision helps the PRM retain its medical foundation during adaptation. 
During policy learning, process feedback and answer rewards target medical reasoning quality in each language, while cross-language semantic alignment encourages transfer of reasoning content between languages.
The alternating training loop uses these signals to improve multilingual medical reasoning (Fig.~\ref{fig:comparison}).

We train multilingual medical reasoning models (\model) and PRMs (\prm) in 8B and 9B variants, across 12 languages: English, Chinese, Spanish, French, Japanese, Korean, Indonesian, Thai, Swahili, Yoruba, Zulu, and Wolof, spanning high-, medium-, and low-resource settings.
We evaluate the resulting models on medical subsets of Global--MMLU and MMLU-ProX for multiple-choice question answering~\cite{singh2025global, xuan2025mmlu}, and on CUREMED-BENCH for open-ended medical question answering~\cite{onyame-etal-2026-cure}.
Our models improve multilingual medical question answering, achieving approximately 3--7.5\% gains over baselines and stronger target language consistency, with further gains from \prm-guided answer selection.
Further analyses and ablation studies highlight the benefits of alternating training and PRM adaptation, particularly for low-resource languages.

In summary, our contributions are as follows:
\begin{itemize}
    \item We introduce an alternating training framework that adapts a medical PRM to evolving policy trajectories and guides multilingual reasoning optimization.
    \item We train \model and \prm in 8B and 9B variants, across 12 languages spanning high-, medium-, and low-resource settings.
    \item We demonstrate gains in multilingual medical reasoning and target-language consistency, with analyses of PRM adaptation, language-specific performance, and key training strategies.
\end{itemize}

\section{Related Work}

\textbf{Process Supervision for Reasoning LLMs.}
Process reward models (PRMs) provide supervision over intermediate reasoning steps, in contrast to outcome reward models that score only the final response.
Early work demonstrated gains from human process supervision and introduced large-scale step annotations for mathematical reasoning~\cite{lightman2024let}, while rollout-based labelling and tree search reduce reliance on manual supervision~\cite{luo2024improve, wang2024math}.
Free Process Reward derives an implicit PRM from an outcome reward model expressed as a likelihood ratio~\cite{yuan2024free}, while Rewarding Progress measures each step's effect on future success~\cite{setlur2025rewarding}.
ProcessBench evaluates reasoning-error detection by PRMs and critics~\cite{zheng2025processbench}.
Medical reasoning introduces additional challenges for process evaluation, as locally plausible reasoning may still lack clinical evidence~\cite{hager2024evaluation, yang2025med, chen2024unlocking}.
Med-PRM~\cite{yun2025med} addresses this problem by verifying intermediate reasoning steps against retrieved clinical guidelines and literature.

\noindent\textbf{Online Adaptation of Reward Models.}
Policy optimization shifts the distribution of generated trajectories, exposing reward-model errors and the risk of overoptimization~\cite{gao2023scaling}.
PRIME trains an implicit PRM online from outcome-labeled policy rollouts and converts token-level likelihood ratios into dense rewards for policy optimization~\cite{cui2025process}.
R-AIRL learns an explicit reasoning critic from both expert demonstrations and policy-generated trajectories, and applies the learned reward to policy training, response reclassification, and error localization~\cite{fanconi2025learning}.
Rewarding Progress similarly uses a separate prover to estimate stepwise gains in future success for online reinforcement learning~\cite{setlur2025rewarding}.
Our framework retains medical step supervision while adapting \prm to current policy trajectories.

\noindent\textbf{Multilingual Medical Reasoning.}
Multilingual corpora and benchmarks have enabled the development and evaluation of medical LLMs across a growing range of languages~\cite{qiu2024towards, labrak2024biomistral}, while gaps between English and non-English performance persist.
Recent work connects English and target-language reasoning through concept-level co-reasoning~\cite{gao-etal-2026-med}, or combine multilingual supervised fine-tuning with GRPO to improve reasoning accuracy~\cite{onyame-etal-2026-cure}.
Parallel translation provides supervision when native step-level annotations are limited.
Translated reasoning data can improve process evaluation in both seen and unseen languages~\cite{wang2025demystifying, wu2024reuse}.
We leverage parallel medical reasoning data to use English reasoning as a reference for other languages. 

\begin{figure*}
    \centering
    \includegraphics[width=\linewidth]{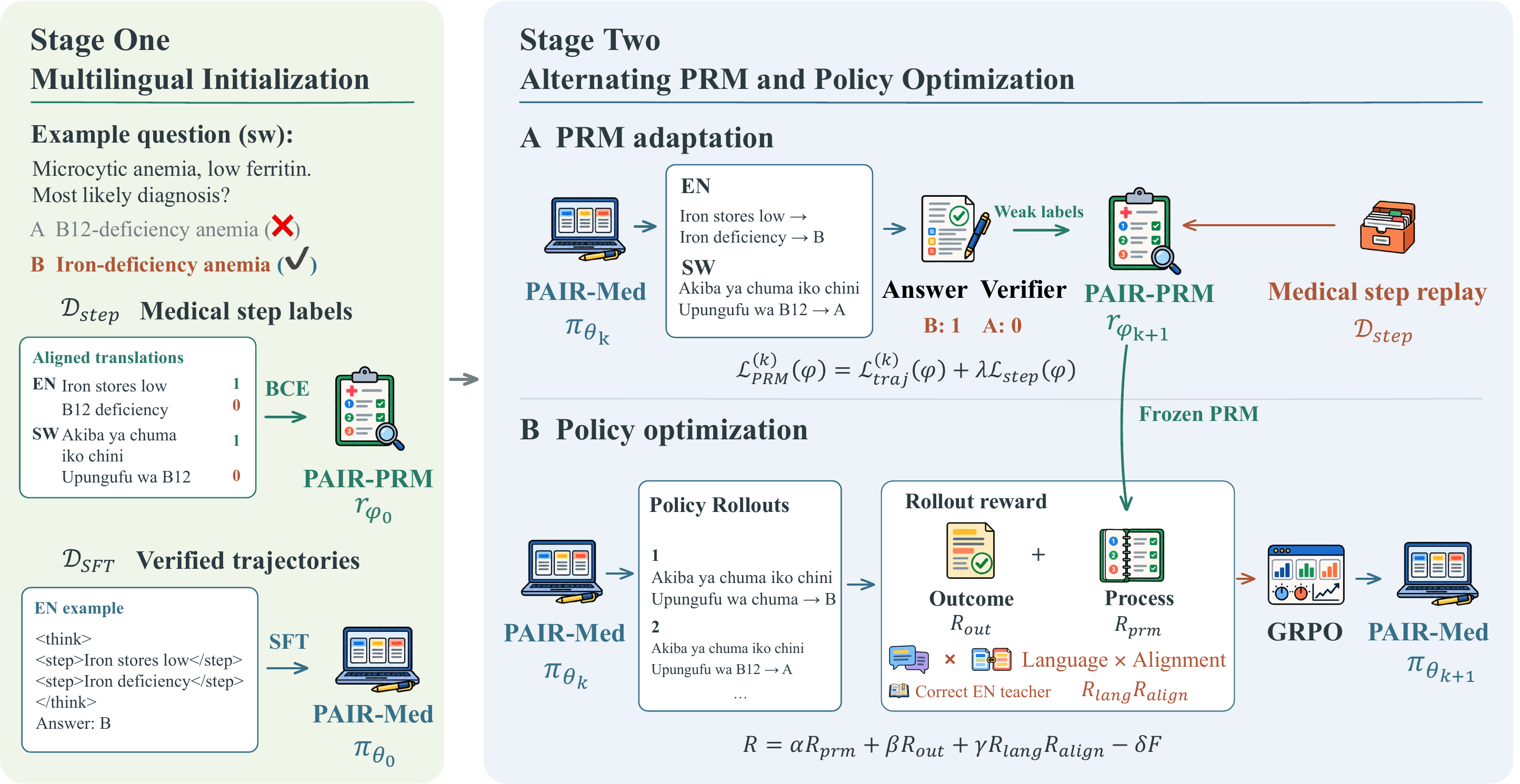}
    \caption{Overview of PAIR-Med: Multilingual Policy and PRM Initialization (Stage One), followed by Alternating PRM Adaptation and Policy Optimization (Stage Two).}
    \label{fig:framework}
\end{figure*}
\section{Method}
\label{sec:method}

We present a two-stage framework for improving multilingual medical reasoning with adaptive process supervision. 
Stage One initializes a shared medical PRM and a multilingual reasoning policy through supervised fine-tuning (SFT). 
Stage Two alternates PRM adaptation and policy optimization, producing \model and \prm (Fig.~\ref{fig:framework}).

\subsection{Stage One: Multilingual Initialization}
\label{sec:offline_initialization}

\noindent\textbf{Policy Cold Start.}
We initialize the reasoning policy using a dataset $\mathcal{D}_{\mathrm{cold}}$ of verified multilingual medical trajectories. 
We format each response with a \texttt{<think>}\ldots\texttt{</think>} block containing the reasoning, and enclose each complete reasoning step in \texttt{<step>}\ldots\texttt{</step>} tags. 
The final answer follows the closing \texttt{</think>} tag. 
We apply standard causal-language-model SFT, computing the loss over response tokens. This cold start teaches the policy to reason in the question language and follow the structured output format, yielding $\pi_{\theta_0}$. 

\textbf{Initializing Process Reward Model.}
The shared PRM uses these explicit steps to score each completed \texttt{<step>}\ldots\texttt{</step>} block. 
It evaluates the step in the context of the question and preceding reasoning steps, producing one scalar score at its closing boundary. 
We initialize the PRM using a separate dataset $\mathcal{D}_{\mathrm{step}}$, constructed from English medical solutions with binary step labels and their step-aligned translations. 
We train it with binary cross-entropy on the step labels across all languages. This parallel supervision establishes the initial multilingual medical PRM $r_{\phi_0}$.

\subsection{Stage Two: Alternating PRM and Policy Optimization}

\noindent\textbf{Generating and Verifying Policy Trajectories.}
At iteration $k$, the current policy $\pi_{\theta_k}$ generates a multilingual rollout set $\mathcal{D}_{k}^{\mathrm{RM}}$ for PRM adaptation. 
For a given question $q^{\ell}$ in language $\ell$, its $j$-th trajectory $\tau_j^{\ell}$ contains reasoning steps and a final answer. 
An answer verifier compares this answer with the ground truth and assigns a binary correctness label $c_j^{\ell}$. 
Verified reference trajectories and correct policy trajectories form the positive set $\mathcal{T}_{k}^{+}$; incorrect policy trajectories form the negative set $\mathcal{T}_{k}^{-}$.


\textbf{Adapting the PRM with Medical Supervision.}
With the policy fixed, we adapt the PRM using trajectory labels as weak step supervision, with replayed medical step supervision. 
Positive and negative trajectories receive labels 1 and 0. 
The binary loss mask $m_t(\tau)$ is 1 if step $t$ contributes to the online loss and 0 otherwise. 
Let $b_t(\phi,\tau)$ denote binary cross-entropy against the trajectory label. 
The masked trajectory loss is
\begin{equation}
    \ell_{\mathrm{traj}}(\phi,\tau)
    = \frac{\sum_t m_t(\tau)\,b_t(\phi,\tau)}
           {\sum_t m_t(\tau)}.
    \label{eq:online_prm_loss}
\end{equation}

The objective $\mathcal{L}_{\mathrm{traj}}^{(k)}$ averages this loss over positive and negative sets with at least one included step.
Alongside trajectory supervision, we replay examples from the fixed step-labelled dataset $\mathcal{D}_{\mathrm{step}}$ used in Stage One. 
Their original step labels provide a medical supervision anchor during each PRM update. 
The combined objective is
\begin{equation}
    \mathcal{L}_{\mathrm{PRM}}^{(k)}(\phi)
    = \mathcal{L}_{\mathrm{traj}}^{(k)}(\phi)
    + \lambda\mathcal{L}_{\mathrm{step}}(\phi),
    \label{eq:combined_prm_loss}
\end{equation}
where $\mathcal{L}_{\mathrm{step}}$ is the binary cross-entropy against the original step labels and $\lambda{=}0.2$ controls its contribution to the PRM update.
Both update the same PRM, starting from $\phi_k$ to obtain $\phi_{k+1}$. 
We then freeze the updated PRM for policy optimization.
\begin{algorithm}[t]
\caption{Alternating PRM and Policy Training}
\label{alg:training}
\small
\begin{algorithmic}[1]
\Require $\mathcal{D}_{\mathrm{step}},\mathcal{D}_{\mathrm{cold}}$; question pool $\mathcal{Q}$; $K,G$
\Ensure $r_{\phi_K},\pi_{\theta_K}$
\State SFT $\pi_{\theta_0}$ on $\mathcal{D}_{\mathrm{cold}}$
\State Train $r_{\phi_0}$ on $\mathcal{D}_{\mathrm{step}}$
\For{$k=0,\ldots,K-1$}
    \State Sample $\mathcal{D}_{k}^{\mathrm{RM}}$ from $\mathcal{Q}$ using $\pi_{\theta_k}$
    \State Verify answers; form $\mathcal{T}_{k}^{+},\mathcal{T}_{k}^{-}$
    \State Update $\phi_k\to\phi_{k+1}$ via Eq.~\ref{eq:combined_prm_loss} ($\pi_{\theta_k}$ fixed)
    \State Freeze $r_{\phi_{k+1}}$; sample $G$ rollouts per prompt
    \State Select English teachers; compute $R$ (Eq.~\ref{eq:combined_policy_reward})
    \State GRPO update $\theta_k\to\theta_{k+1}$; reference $\pi_{\theta_0}$
\EndFor
\State \textbf{return} $r_{\phi_K},\pi_{\theta_K}$
\end{algorithmic}
\end{algorithm}
\textbf{Scoring Outcomes and Reasoning Steps.}
The policy samples a group of $G$ candidate trajectories for each question--language pair.
We assign a reward to each complete rollout $\tau$. 
The outcome reward $R_{\mathrm{out}}(\tau)$ is one when its final answer is correct and zero otherwise. 
The process reward $R_{\mathrm{prm}}(\tau)$ is the mean scores over its parsed reasoning steps.

\textbf{Rewarding Target-Language Use.}
The language reward $R_{\mathrm{lang}}$ measures adherence to the requested language in the parsed reasoning steps. 
We use a character n-gram classifier fitted on training prompts and their language labels, and average the predicted target-language probabilities with weights proportional to the number of letters in each text segment. 
Appendix~\ref{sec:appendix_language_consistency} describes the classifier, text preprocessing, and reward computation.

\textbf{Aligning with an English Teacher.}
We use English as the teacher language to transfer the model's stronger English reasoning capabilities to other languages. 
We align each target-language rollout with a correct English teacher generated for the same question, using the frozen multilingual Granite Embedding model~\cite{awasthy2025granite} to encode their reasoning steps.
Let $S_{\mathrm{tgt}}$ average each target step's highest cosine similarity to an English step, and let $S_{\mathrm{en}}$ be the reverse-direction average. 
The alignment reward is
\begin{equation}
    R_{\mathrm{align}}
    = \operatorname{scale}\!\left(
      \frac{S_{\mathrm{tgt}}+S_{\mathrm{en}}}{2}\right).
    \label{eq:alignment_reward}
\end{equation}
Here, $\operatorname{scale}$ maps the mean similarity to $[0,1]$. Appendix~\ref{sec:appendix_semantic_alignment} provides the full score definition, teacher selection, and encoding details.

\textbf{Combining Rollout Rewards.}
For a rollout, the overall reward is
\begin{equation}
\begin{aligned}
    R
    &= \alpha R_{\mathrm{prm}}+\beta R_{\mathrm{out}} \\
    &\quad+\gamma R_{\mathrm{lang}}R_{\mathrm{align}}-\delta F,
\end{aligned}
    \label{eq:combined_policy_reward}
\end{equation}
where $F$ is one for an invalid output format and zero otherwise. 
The default configuration uses process and outcome weights $\alpha=\beta=1$, a joint language reward weight $\gamma=0.3$, and a format penalty weight $\delta=0.1$. 
The entire language-related term is zero for English rollouts or when no eligible English teacher is available. 
A target-language rollout can receive this bonus even when its final answer is incorrect. 

\textbf{Optimizing the Reasoning Policy.}
We optimize the policy with Group Relative Policy Optimization (GRPO) \citep{shao2024deepseekmath}. 
Rewards are normalized within the $G$ trajectories sampled for the same question and language to obtain group advantages. 
We then maximize the GRPO objective with a KL penalty to the fixed SFT reference policy. 
The resulting policy $\pi_{\theta_{k+1}}$ generates new trajectories for PRM adaptation in the next iteration. 
Algorithm~\ref{alg:training} summarizes both training stages.

\section{Experiments}
\label{sec:experiments}

\subsection{Experimental Setups}
\paragraph{Backbones and Languages.}
We select two recent open-weight reasoning models, Qwen3-8B~\cite{yang2025qwen3} and Qwen3.5-9B~\cite{team2026qwen3} as our backbones.
Although both models demonstrate strong overall reasoning capabilities, their multilingual performance remains imbalanced, with noticeably better results in high-resource languages.
To study whether our method can improve medical reasoning across languages with different levels of resource availability, we construct a multilingual training set covering three resource groups.
Specifically, we include English (en), Spanish (es), French (fr), Chinese (zh), and Japanese (ja) as high-resource languages; 
Korean (ko), Thai (th), and Indonesian (id) as medium-resource languages; and Swahili (sw), Yoruba (yo), Wolof(wo) and Zulu (zu) as low-resource languages.
To further examine cross-lingual generalization, we additionally select Hausa (ha), Amharic (am), Vietnamese (vi), Hindi (hi), Turkish (tr), and Bengali (bn) for out-of-domain (OOD) evaluation.

\paragraph{Training data.}
We conducted training based on the Med-PRM training set~\cite{yun2025med}, which contains 1,0644 medical questions. 
Each question is associated with multiple CoT reasoning trajectories and medical guideline-grounded process rewards.
To construct the multilingual training data, we use GPT-5.5-2026-04-23~\cite{openai2026gpt55} to translate the original English questions and CoT responses into es/fr/zh/ja/ko/th/id/sw/yo/zu.

The parallel multilingual data is used in two stages.
In Stage One, we use 1K parallel question--answer examples for SFT as a cold start for the policy model, and 6K parallel questions with process-level reward annotations for PRM training.
In Stage Two, the remaining samples are used as a shared data pool for GRPO-based policy training and iterative PRM updates.
At this stage, the policy and PRM are trained on disjoint question subsets to avoid overlap within each iteration.

\paragraph{Benchmarks and Metrics.}
For multilingual medical reasoning evaluation, we consider the medical category of Global-MMLU~\cite{singh2025global} and the health category of MMLU-ProX~\cite{xuan2025mmlu}, which contain 1,505 and 687 parallel multiple-choice questions.
Since Global-MMLU does not provide Wolof and Zulu versions, we use GPT-6-sol~\cite{openai2026gpt6sol} to translate the corresponding questions into these two languages.
For open-ended question answer and OOD evaluation, we use CUREMed-BENCH~\cite{onyame-etal-2026-cure}, which includes 1,431 open-ended questions in 13 languages. 
CUREMED-BENCH contains language-specific questions.

For Global-MMLU and MMLU-ProX, we report answer accuracy as the evaluation metric.
For CUREMED-BENCH, we evaluate logical accuracy (LA) and language consistency (LC) according to the official evaluation protocol. 
LA measures answer correctness, while LC assesses whether both reasoning and answers use the target language. 
Following ~\cite{zhang2026think}, we report LA\&LC measuring the proportion of responses satisfying both correctness and consistency, with separate LA and LC scores in the Appendix~\ref{sec:appendix_curemed_details}.
All metrics are evaluated using \texttt{GPT-6.1-sol}~\citep{openai2026gpt61sol}.

\paragraph{Baselines.}
We compare against general-purpose reasoning models and medical LLMs of comparable or larger scales. 
General-purpose baselines include Qwen3-8B and Qwen3.5-9B. At the 7--8B scale, we include Apollo2-7B~\cite{zheng2025efficiently}, BioMistral-7B~\cite{labrak2024biomistral}, CURE-Med-7B~\cite{onyame-etal-2026-cure}, and HuatuoGPT-3-8B~\cite{huatuogpt3}. 
We also include larger medical models: MedGemma-27B~\cite{sellergren2025medgemma}, CURE-Med-32B~\cite{onyame-etal-2026-cure}, HuatuoGPT-3-32B~\cite{sellergren2025medgemma}, and Baichuan-M2-32B~\cite{dou2025baichuan}.

\paragraph{Implementation Details.}
We set the training iterations for alternating training to 3. The detailed training settings and generation configurations are listed in the Appendix~\ref{sec:appendix_implementation}.

\begin{table*}[t]
\centering
\small
\setlength{\tabcolsep}{4.5pt}
\resizebox{\textwidth}{!}{
\begin{tabular}{l*{4}{c}@{\hspace{1em}}*{4}{c}}
\toprule
& \multicolumn{4}{c}{\textbf{Global-MMLU}} & \multicolumn{4}{c}{\textbf{MMLU-ProX}} \\
\cmidrule(lr){2-5}\cmidrule(lr){6-9}
Model & \textbf{High} & \textbf{Medium} & \textbf{Low} & \textbf{Avg.} & \textbf{High} & \textbf{Medium} & \textbf{Low} & \textbf{Avg.} \\
\midrule
Qwen3-8B~\cite{yang2025qwen3} & 61.95 & 62.27 & 32.19 & 52.11 & 43.81 & 39.32 & 18.81 & 34.35 \\
Apollo2-7B~\cite{zheng2025efficiently} & 62.38 & 51.70 & 31.67 & 49.47 & 38.86 & 28.57 & 11.75 & 27.25 \\
BioMistral-7B~\cite{labrak2024biomistral} & 41.19 & 39.38 & 28.27 & 36.43 & 18.54 & 10.24 & 13.32 & 14.72 \\
CURE-Med-7B~\cite{onyame-etal-2026-cure} & 62.62 & 54.32 & 33.18 & 50.73 & 33.82 & 27.17 & 14.96 & 25.87 \\
HuatuoGPT-3-8B~\cite{huatuogpt3} & 70.23 & 63.19 & 34.05 & 56.41 & 46.81 & 37.84 & 19.39 & 35.43 \\
\rowcolor{LightGreen} 
\textbf{\model-8B}  & \underline{70.65} & \underline{65.66} & \underline{41.19} & \underline{59.58} & \underline{49.91} & \underline{41.41} & \underline{22.51} & \underline{38.65} \\
\rowcolor{LightGreen}
\textbf{\model-8B w/ \prm-8B} & \textbf{76.41} & \textbf{71.71} & \textbf{45.66} & \textbf{64.99} & \textbf{56.36} & \textbf{49.10} & \textbf{27.15} & \textbf{44.81} \\
\midrule
Qwen3.5-9B~\cite{team2026qwen3} & 76.24 & 68.78 & 53.32 & 66.73 & 60.06 & 49.69 & 32.75 & 48.36 \\
MedGemma-27B~\cite{sellergren2025medgemma} & 60.59 & 65.43 & 46.40 & 57.07 & 56.18 & 50.80 & 30.84 & 46.39 \\
CURE-Med-32B~\cite{onyame-etal-2026-cure} & 77.74 & 72.87 & 43.98 & 65.27 & 60.87 & 52.20 & 24.16 & 46.47 \\
HuatuoGPT-3-32B~\cite{huatuogpt3} & 78.80 & \underline{74.89} & 42.59 & 65.75 & 62.15 & \underline{55.69} & 23.43 & 47.63 \\
Baichuan-M2-32B~\cite{dou2025baichuan} & 78.13 & 74.02 & 40.84 & 64.67 & 62.32 & 54.14 & 21.94 & 46.82 \\
\rowcolor{LightGreen}
\textbf{\model-9B} & \underline{78.92} & 72.04 & \underline{56.53} & \underline{69.74} & \underline{63.90} & 54.78 & \underline{36.20} & \underline{52.39} \\
\rowcolor{LightGreen}
\textbf{\model-9B w/ \prm-9B} & \textbf{81.13} & \textbf{75.73} & \textbf{62.60} & \textbf{73.60} & \textbf{66.38} & \textbf{58.47} & \textbf{42.50} & \textbf{56.44} \\
\bottomrule
\end{tabular}
}
\caption{Accuracy (\%) on Global-MMLU-medical and MMLU-ProX-health. High, Medium, and Low denote language resource groups. Within each model group, the highest and second-highest values per column are bolded and underlined, respectively. \model accuracy is averaged over $N{=}10$ samples, while w/ \prm selects the highest-scoring response among $N{=}10$ candidates. Per-language results are provided in the Appendix.~\ref{sec:appendix_multilingual_results}.}
\label{tab:multilingual_results}
\end{table*}

\begin{table*}[t]
\centering
\small
\setlength{\tabcolsep}{2.8pt}
\resizebox{\textwidth}{!}{
\begin{tabular}{l*{8}{c}@{\hspace{0.8em}}*{7}{c}}
\toprule
& \multicolumn{8}{c}{\textbf{In-Domain}} & \multicolumn{7}{c}{\textbf{Out-of-Domain}} \\
\cmidrule(lr){2-9}\cmidrule(lr){10-16}
Model & \textbf{es} & \textbf{fr} & \textbf{ja} & \textbf{ko} & \textbf{th} & \textbf{sw} & \textbf{yo} & \textbf{Avg.} & \textbf{vi} & \textbf{hi} & \textbf{tr} & \textbf{am} & \textbf{bn} & \textbf{ha} & \textbf{Avg.} \\
\midrule

Qwen3-8B & 0.00 & 0.00 & 0.00 & 0.00 & 0.00 & 0.00 & 0.00 & 0.00 & 0.00 & 0.00 & 0.00 & 0.00 & 0.00 & 0.00 & 0.00 \\
HuatuoGPT-3-8B & 0.00 & 0.00 & 0.00 & 0.00 & 0.00 & 0.00 & 0.00 & 0.00 & 0.00 & 0.00 & 0.00 & 0.00 & 0.00 & 0.00 & 0.00 \\
Qwen3.5-9B & 0.00 & 0.00 & 0.00 & 0.00 & 0.00 & 0.00 & 0.00 & 0.00 & 0.00 & 0.00 & 0.00 & 0.00 & 0.00 & 0.00 & 0.00 \\
CURE-Med-7B & 11.43 & 8.57 & 4.76 & 4.00 & 2.04 & 0.00 & 0.00 & 4.40 & 6.67 & 0.75 & 1.79 & 0.00 & 0.00 & 0.00 & 1.53 \\
\rowcolor{LightGreen}
\textbf{\model-8B} & 22.86 & 25.71 & 15.24 & \textbf{13.33} & \textbf{9.18} & 2.86 & 0.00 & 12.74 & 14.29 & 6.72 & 8.04 & 0.00 & 0.71 & 0.00 & 4.96 \\
\rowcolor{LightGreen}
\textbf{\model-9B} & \textbf{38.10} & \textbf{37.14} & \textbf{18.10} & 8.00 & 7.14 & \textbf{9.29} & \textbf{3.23} & \textbf{17.28} & \textbf{22.86} & \textbf{14.18} & \textbf{12.50} & \textbf{0.95} & \textbf{2.14} & \textbf{1.27} & \textbf{8.98} \\
\bottomrule
\end{tabular}
}
\caption{Zero-shot prompting results on CUREMED-BENCH: LC\&LA. Reasoning baselines use thinking mode.}
\label{tab:curemed_bench_results}
\end{table*}

\subsection{Main Results}
\textbf{\model improves multilingual reasoning performance.}
As shown in Table~\ref{tab:multilingual_results}, \model outperforms its corresponding Qwen backbone across all resource groups on both benchmarks. \model-8B achieves overall accuracies of 59.58\% on Global-MMLU and 38.65\% on MMLU-ProX, compared with 52.11\% and 34.35\% for Qwen3-8B.
Similarly, \model-9B reaches 69.74\% and 52.39\%, compared with 66.73\% and 48.36\% for Qwen3.5-9B.
\model-9B also achieves higher overall accuracy than the evaluated 27--32B medical models on both benchmarks.

\textbf{\prm provides additional improvements in reasoning quality.}
Introducing \prm improves \model at both model scales and across all resource groups. For \model-8B, overall accuracy increases by 5.41 points on Global-MMLU and 6.16 points on MMLU-ProX, reaching 64.99\% and 44.81\%.
\model-9B gains an additional 3.86 and 4.05 points, reaching 73.60\% and 56.44\%. 
These results show that PRM contributes additional accuracy gains.

\textbf{Improvements extend to low-resource languages.}
On Global-MMLU, \model-9B improves low-resource accuracy from 53.32\% to 56.53\% over Qwen3.5-9B, and further to 62.60\% with PRM. 
On MMLU-ProX, the corresponding scores increase to 42.50\%. 
With PRM, the high-to-low resource accuracy gap decreases from 22.92 to 18.53 points on Global-MMLU and from 27.31 to 23.88 points on MMLU-ProX.

\textbf{Multilingual medical QA reveals a trade-off between language consistency and accuracy.}
Under zero-shot prompting, \model-8B achieves in-domain/out-of-domain LA\&LC averages of 12.74\%/4.96\%, while \model-9B reaches 17.28\%/8.98\%, outperforming CURE-Med-7B in both domains (Table~\ref{tab:curemed_bench_results}). 
The Qwen backbones and HuatuoGPT-3-8B score zero, as they generally reason in English regardless of the input language. However, separate results (Appendix~\ref{sec:appendix_curemed_details}) show that higher language consistency is accompanied by lower logical accuracy. 
For example, \model-9B achieves 76.52\% LC but 48.77\% LA out-of-domain, compared with 68.02\% LA for Qwen3.5-9B. 
These findings highlight the need to improve both language consistency and reasoning accuracy.


\section{Analysis}
\label{sec:analysis}

We address three research questions: (1) RQ1: How does alternating training affect the reasoning policy? (2) RQ2: How does alternating training affect \prm? (3) RQ3: How does reasoning quality vary across languages, and how effectively does \prm assess it?

\subsection{RQ1: Policy Performance}
\label{sec:training_method_analysis}

We compare six training configurations: the base model, Stage-One SFT cold start, Native SFT, Native GRPO, frozen-PRM training, and our alternating framework (\model). 
Native SFT uses all training data, while Native GRPO starts from the Stage-One SFT model and uses the same outcome ($R_{out}$) and language ($R_{lang}$) rewards. Frozen-PRM training uses the fixed Stage-One PRM for three policy optimization rounds. 
Figure~\ref{fig:training_method_gains} shows accuracy changes relative to the base models, with detailed results in Appendix~\ref{sec:appendix_training_methods}.

\textbf{Alternating training improves policy accuracy across resource levels.}
\model achieves the highest overall accuracy at both model sizes on both benchmarks. 
At 8B, fixed-PRM training has slightly higher high-resource accuracy on MMLU-ProX, while alternating training performs better in the medium- and low-resource groups. 
Its advantage over the frozen scorer is more pronounced for low-resource languages at 8B: 41.19\% versus 32.00\% on Global-MMLU and 22.51\% versus 16.55\% on MMLU-ProX.

\textbf{Multilingual initialization changes the reasoning language and format.}
Base models generally reason in English, even for non-English questions, while Stage-One SFT encourages explicit reasoning in the input language.
However, accuracy gains are uneven: at 8B, only high-resource accuracy improves, while at 9B, accuracy decreases across all groups. 
Stage Two subsequently optimizes the initialized policy using process, outcome, and language-related rewards. 
The final \model policies recover the cold-start losses and exceed their original backbones consistently.

\textbf{Low-resource performance highlights the role of language supervision.}
Native GRPO exceeds Native SFT in high- and medium-resource accuracy but yields lower Low-resource accuracy in all four settings. 
Consistent with prior work on multilingual data and supervised initialization~\cite{qiu2024towards, onyame-etal-2026-cure}, these results suggest the importance of sufficient Low-resource supervision and target-language reasoning capabilities before further optimization.
\model incorporates multilingual initialization with process- and language-related feedback, achieving higher accuracy than both baselines across all resource groups.

\begin{figure*}[t]
\centering
\includegraphics[width=\linewidth]{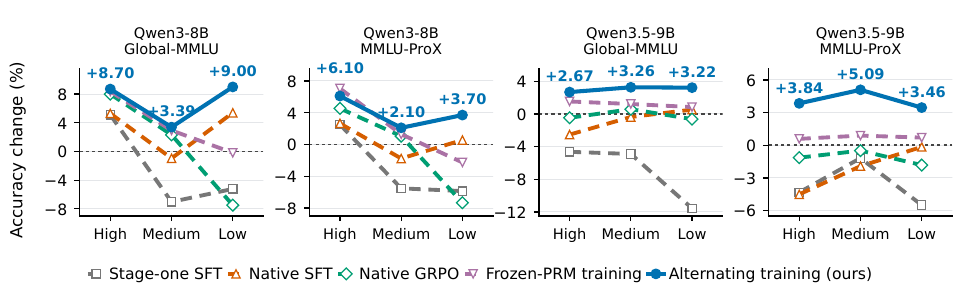}
\caption{Accuracy changes relative to each base model across high-, medium-, and low-resource languages.}
\label{fig:training_method_gains}
\end{figure*}

\subsection{RQ2: PRM Adaptation}
\label{sec:prm_stage_analysis}

\begin{figure}
\centering
\includegraphics[width=\linewidth]{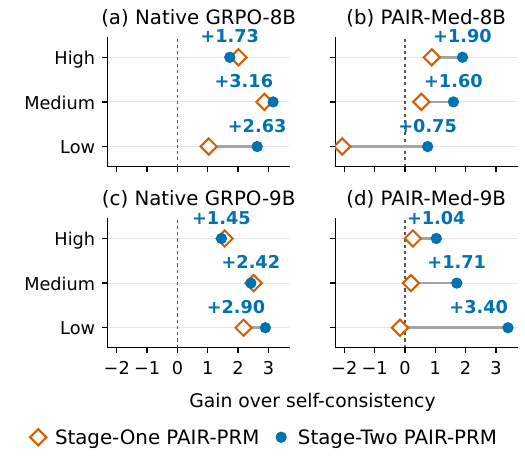}
\caption{Stage-One and Stage-Two PRM best-of-$N$ answer selection on Global-MMLU. The vertical zero line denotes self-consistency accuracy. }
\label{fig:prm_stage_comparison}
\end{figure}

We assess PRM adaptation on Global-MMLU. 
We compare the initial Stage-One PRM with the Stage-Two PRM adapted during alternating training, using candidates generated by Native GRPO and \model at both model sizes.
For each question, we sample 10 candidate responses.
Self-consistency selects an answer by majority vote, while PRM-assisted inference uses best-of-$N$ selection. 
Figure~\ref{fig:prm_stage_comparison} reports resource-group accuracy relative to self-consistency. 
Detailed scores are given in the Appendix~\ref{sec:appendix_prm_stages}.

\textbf{The adapted PRM improves answer selection on \model.}
The Stage-Two PRM exceeds the Stage-One PRM in all three resource groups, with the largest improvements in Low-resource languages. 
For \model-9B, Low-resource accuracy reaches 62.60\% with Stage Two, compared with 59.04\% with Stage One and 59.20\% with self-consistency. 
At both sizes, the initial scorer falls below self-consistency in the Low-resource group, whereas the adapted scorer exceeds self-consistency in all three groups.

\textbf{The adapted PRM generalizes to Native GRPO outputs.}
The Stage-Two PRM remains effective on Native GRPO trajectories after adaptation within the alternating framework. 
Its best-of-$N$ answer selection exceeds majority-vote self-consistency in all three resource groups at both model sizes, extending its utility beyond \model. 
Compared with Stage One, it also achieves higher overall and Low-resource accuracy at both sizes, with small decreases in High-resource accuracy and in Medium-resource accuracy at 9B.

\subsection{RQ3: Reasoning and PRM Quality}
\label{sec:reasoning_quality_analysis}

We assess multilingual reasoning quality and the PRM's ability to evaluate it on Global-MMLU and MMLU-ProX. 
For each \model policy, we sample 50 responses per language from each benchmark. We use \texttt{GPT-6.1-sol} to rate each reasoning step for medical correctness and clinical or scientific logic on $[0,1]$, alongside the corresponding Stage-Two PRM.
We average step scores within each response and compute Spearman's rank correlation between GPT judge and PRM scores at the response level.
Table~\ref{tab:reasoning_quality_groups} summarizes the results, and Appendix~\ref{sec:appendix_reasoning_quality} provides the detailed results.

\textbf{Reasoning quality varies across language resource levels.}
GPT judge ratings decrease from High- to Low-resource languages for both model sizes. 
\model-9B receives higher mean ratings than \model-8B across all three resource groups. 
However, Low-resource languages receive the lowest ratings for both models, with mean scores of $0.353$ and $0.501$ at 8B and 9B.

\textbf{PRM scores show positive rank correlation with judge ratings.}
Both policy--PRM pairs show positive correlations with GPT judge ratings across all languages, with overall mean Spearman correlations of $0.321$ for 8B and $0.415$ for 9B. Correlations are lowest in Low-resource languages for both pairs, at $0.213$ and $0.363$, respectively. 
The 9B pair also shows higher correlations than the 8B pair in all three resource groups, with larger differences in Medium- and Low-resource languages.
In addition, PRM scores vary less across resource groups than GPT judge ratings, suggesting that the PRMs may not fully capture the lower reasoning quality observed in Low-resource languages.

\begin{table}[t]
\centering
\small
\setlength{\tabcolsep}{2.0pt}
\begin{tabular*}{\linewidth}{@{\extracolsep{\fill}}cccc@{\hspace{6pt}\extracolsep{\fill}}ccc@{}}
\toprule
& \multicolumn{3}{c}{\textbf{\model-8B}} & \multicolumn{3}{c}{\textbf{\model-9B}} \\
\cmidrule(lr){2-4}\cmidrule(lr){5-7}
Group & GPT & PRM-8B & $\rho$ & GPT & PRM-9B & $\rho$ \\
\midrule
High & 0.748 & 0.705 & 0.387 & 0.801 & 0.719 & 0.403 \\
Medium & 0.652 & 0.662 & 0.356 & 0.731 & 0.667 & 0.506 \\
Low & 0.353 & 0.655 & 0.213 & 0.501 & 0.617 & 0.363 \\
\midrule
Avg. & 0.592 & 0.677 & 0.321 & 0.684 & 0.672 & 0.415 \\
\bottomrule
\end{tabular*}
\caption{Reasoning quality and PRM agreement. GPT denotes LLM-as-the-judge validity rating.}
\label{tab:reasoning_quality_groups}
\end{table}
\section{Conclusion}

We presented a policy--reward alternating training framework in which a shared multilingual medical PRM adapts to current policy trajectories and guides subsequent reasoning-policy updates. 
The framework combines replayed medical step supervision with process, outcome, and cross-language rewards. 
Across 12 languages, \model improves medical reasoning accuracy over its corresponding backbones on Global-MMLU and MMLU-ProX. 
Comparisons with frozen-PRM training support the benefit of alternating updates, while the adapted PRM further improves answer selection, especially in low-resource languages. 
On CUREMED-BENCH, \model improves target-language consistency. 
These findings show the value of adapting process feedback as the multilingual reasoning policy evolves during training.

\section*{Limitations}

First, Stage-Two \prm adaptation uses final-answer correctness as weak step supervision, which can mislabel individual reasoning steps. Stage-One supervision uses translated English medical trajectories, and cross-language alignment relies on English teachers. Native-language step annotations and locally grounded clinical examples would provide more direct process supervision and broaden its linguistic and clinical coverage. Second, we test the framework on Qwen backbones at 8B and 9B using medical QA benchmarks. In CUREMED-BENCH, \model achieves higher language consistency but lower average logical accuracy than its corresponding backbones in both in-domain and out-of-domain languages, indicating uneven transfer to open-ended tasks. Reasoning-quality assessment also relies on a single LLM judge. Multilingual clinician review would strengthen the assessment of reasoning validity and clinical applicability. Third, repeated rollout generation and alternating policy--PRM updates add to training time, which can limit scalability under fixed computational budgets. Finally, our multilingual evaluation relies on an LLM judge; review by physicians proficient in the evaluated languages would strengthen the assessment.

\section*{Ethics Statement}

This study uses existing medical question-answering datasets and model-generated reasoning trajectories. \model{} and \prm{} are intended for research; clinical use requires independent validation and qualified medical oversight. 
Responsible use should account for performance disparities across languages and potential biases associated with translated supervision and English alignment.



\bibliography{reference}
\clearpage
\appendix

\label{sec:appendix}

\section{Implementation Details}
\label{sec:appendix_implementation}



\subsection{Training Configuration}
\label{sec:appendix_shared_training}

All training stages use LoRA with rank 16, scaling parameter $\alpha_{\mathrm{LoRA}}=32$, and dropout 0.05. 
The common settings are BF16 precision, gradient checkpointing, a cosine learning-rate schedule, a warmup ratio of 0.03, zero weight decay, gradient clipping at 1.0, and random seed 42. 
The maximum training sequence length is 4,096 tokens for both policies and PRMs. Stage-One training uses distributed data parallelism for 8B/9B models.

\subsection{Stage-One Initialization}
\label{sec:appendix_stage1_implementation}


\begin{table}[t]
\centering
\small
\setlength{\tabcolsep}{4pt}
\begin{tabular}{lcc}
\toprule
Setting & SFT Cold-start &  PRM Initialization \\
\midrule
Training epochs & 3 & 1 \\
Learning rate & $2\times10^{-5}$ & $2\times10^{-5}$ \\
Global batch size & 64 & 64 \\
Training GPUs & 8 & 8 \\
\bottomrule
\end{tabular}
\caption{Stage One optimization settings.}
\label{tab:stage1_hyperparameters}
\end{table}
Table~\ref{tab:stage1_hyperparameters} summarizes the SFT cold-start and multilingual PRM initialization settings. 
Cold-start SFT enables the thinking chat template and uses the response format, with the answer following the reasoning block. 
Prompt tokens are masked from the loss, and sequence packing is disabled. 
Each GPU processes one example per microbatch with eight gradient-accumulation steps, giving a global batch size of $8\times1\times8=64$.
The PRM predicts a scalar at the final token of each complete \texttt{</step>} marker. 
One causal forward pass scores all completed steps in a trajectory. 
We apply binary cross-entropy with logits, first averaging over the steps within each trajectory and then over trajectories. 

\subsection{Stage-Two Alternating Training}
\label{sec:appendix_stage2_implementation}
Each run starts from a Stage-One cold-start policy and a Stage-One PRM of the same backbone size. 
Each round follows the generation, verification, PRM adaptation, rescoring, and GRPO update sequence in Algorithm~\ref{alg:training}, with one epoch for each PRM and policy update. 
PRM adaptation and policy optimization use two independent sampling streams, each containing 128 complete parallel question groups per round. 
Each stream therefore contains $128\times12=1,536$ prompts. 
Sampling four candidates per prompt produces 6,144 trajectories per stream, or 12,288 candidates across the two streams per round. 

\begin{table}[t]
\centering
\small
\setlength{\tabcolsep}{4pt}
\begin{tabular}{lc}
\toprule
Setting & Value \\
\midrule
Sampling temperature & 0.8 \\
Top-$p$ / top-$k$ & 0.95 / 50 \\
Maximum new tokens & 2,048 \\
PRM / policy learning rate & $10^{-5}$ / $5\times10^{-6}$ \\
PRM / policy global batch size & 8 / 8 \\
GRPO clipping parameter $\epsilon$ & 0.2 \\
KL coefficient & 0.01 \\
Offline PRM trajectories & 128 \\
\bottomrule
\end{tabular}
\caption{Generation and optimization settings for the alternating-training runs. Each PRM and policy update uses one epoch.}
\label{tab:stage2_hyperparameters}
\end{table}



\subsection{Computation}
\label{sec:appendix_compute}

We train on the eight NVIDIA A100 GPUs with 40\,GB memory each. 
Table~\ref{tab:stage1_compute} reports Stage One training runtimes and GPU-hours.
The Stage Two figures in Table~\ref{tab:stage2_compute} include PRM and policy alternating training. 

\begin{table*}[t]
\centering
\small
\setlength{\tabcolsep}{7pt}
\begin{tabular}{lrrrr}
\toprule
& \multicolumn{2}{c}{SFT Cold Start} & \multicolumn{2}{c}{Multilingual PRM Initialization} \\
\cmidrule(lr){2-3}\cmidrule(lr){4-5}
Backbone & Training hours & GPU-hours & Training hours & GPU-hours \\
\midrule
Qwen3-8B & 0.73 & 5.86 & 13.39 & 107.10 \\
Qwen3.5-9B & 2.38 & 19.02 & 45.88 & 367.07 \\
\bottomrule
\end{tabular}
\caption{Stage-One training cost on eight GPUs.}
\label{tab:stage1_compute}
\end{table*}

\begin{table}[t]
\centering
\small
\setlength{\tabcolsep}{4pt}
\begin{tabular}{lc}
\toprule
Backbone & Alternating Training \\
\midrule
Qwen3-8B & $7.52$  \\
Qwen3.5-9B & $25.22$ \\
\bottomrule
\end{tabular}
\caption{GPU-hours for PRM and policy updates over all rounds of Stage Two training.}
\label{tab:stage2_compute}
\end{table}

\subsection{API Cost}
\label{sec:appendix_translation_cost}

We translate the English datasets into 11 non-English training languages using \texttt{GPT-5.5}. 
The Med-PRM corpus contains approximately 10,644 questions per language, while the multilingual PRM corpus comprises 6,000 questions and 67,391 solutions per language, totaling 741,301 translated solutions across all 11 languages. 
We estimate the translation cost at approximately US\$1,000 per language (US\$11,000 in total), with an additional US\$200 for evaluation using \texttt{GPT-6.1-sol} as the LLM judge. 
The total estimated API cost is therefore approximately US\$11,200.

\section{Multilingual Results by Language}
\label{sec:appendix_multilingual_results}

Tables~\ref{tab:global_mmlu_results} and~\ref{tab:mmluprox_results} report the detailed results underlying Table~\ref{tab:multilingual_results}. 
Languages are grouped into high-resource (en, es, fr, zh, and ja), mid-resource (ko, id, and th), and low-resource (sw, wo, yo, and zu) groups. 
Each table contains individual language scores. 
The w/ \prm rows list the corresponding with-\prm configurations of \model-8B and \model-9B, using the \prm variant of the same parameter size in each case. 

\begin{table*}[t]
\centering
\small
\setlength{\tabcolsep}{3.2pt}
\resizebox{\textwidth}{!}{
\begin{tabular}{l*{13}{c}}
\toprule
& \multicolumn{5}{c}{High} & \multicolumn{3}{c}{Medium} & \multicolumn{4}{c}{Low} & \\
\cmidrule(lr){2-6}\cmidrule(lr){7-9}\cmidrule(lr){10-13}
Model & en & es & fr & zh & ja & ko & id & th & sw & wo & yo & zu & Avg. \\
\midrule
Qwen3-8B & 68.04 & 64.78 & 60.79 & 62.65 & 53.48 & 54.46 & 65.54 & \underline{66.80} & 37.54 & 34.60 & 25.90 & 30.71 & 52.11 \\
Apollo2-7B & 69.90 & 57.81 & 57.34 & 66.84 & 60.01 & 52.62 & 54.41 & 48.07 & 38.67 & 29.32 & 29.36 & 29.31 & 49.47 \\
BioMistral-7B & 55.08 & 43.78 & 45.91 & 37.67 & 23.52 & 31.56 & 37.24 & 49.33 & 30.36 & 29.79 & 28.84 & 24.09 & 36.43 \\
CURE-Med-7B & 69.23 & 60.53 & 61.46 & 64.25 & 57.61 & 53.28 & 57.94 & 51.73 & 39.13 & 30.72 & 32.82 & 30.04 & 50.73 \\
HuatuoGPT-3-8B & \underline{75.81} & 70.09 & 69.30 & 70.23 & \underline{65.71} & 57.34 & 68.62 & 63.60 & 38.40 & 35.53 & 30.96 & 31.30 & 56.41 \\
\rowcolor{LightGreen}
\textbf{\model-8B} & 75.68 & \underline{70.24} & \underline{70.33} & \underline{71.42} & 65.58 & \underline{61.46} & \underline{70.39} & 65.13 & \underline{44.72} & \underline{44.62} & \underline{37.45} & \underline{37.97} & \underline{59.58} \\
\rowcolor{LightGreen}
\textbf{\model-8B w/ \prm} & \textbf{80.33} & \textbf{75.81} & \textbf{75.88} & \textbf{77.54} & \textbf{72.49} & \textbf{69.17} & \textbf{76.43} & \textbf{69.53} & \textbf{48.30} & \textbf{47.96} & \textbf{42.25} & \textbf{44.13} & \textbf{64.99} \\
\midrule
Qwen3.5-9B & 80.99 & 76.81 & 76.27 & 76.07 & 71.08 & 67.83 & 77.10 & 61.40 & 60.33 & 51.17 & 48.42 & 53.34 & 66.73 \\
MedGemma-27B & 72.02 & 50.03 & 47.24 & 62.90 & 70.76 & 59.80 & 70.29 & 66.20 & 50.36 & 32.20 & 44.70 & \underline{58.34} & 57.07 \\
CURE-Med-32B & 80.93 & 77.54 & 77.54 & 77.87 & 74.81 & 68.77 & 76.43 & 73.40 & 51.76 & 41.95 & 39.53 & 42.66 & 65.27 \\
HuatuoGPT-3-32B & \underline{82.92} & 78.53 & 79.13 & 77.67 & \underline{75.75} & 71.56 & 77.57 & \textbf{75.53} & 54.09 & 37.07 & 36.87 & 42.32 & 65.75 \\
Baichuan-M2-32B & 82.39 & 78.01 & 76.87 & 78.41 & 74.95 & 71.42 & 76.70 & \underline{73.93} & 47.51 & 38.07 & 38.80 & 38.98 & 64.67 \\
\rowcolor{LightGreen}
\textbf{\model-9B} & 82.67 & \underline{79.46} & \underline{79.53} & \underline{78.65} & 74.28 & \underline{72.19} & \underline{78.93} & 64.99 & \underline{61.71} & \underline{53.05} & \underline{53.10} & 58.27 & \underline{69.74} \\
\rowcolor{LightGreen}
\textbf{\model-9B w/ \prm} & \textbf{84.32} & \textbf{81.59} & \textbf{81.99} & \textbf{80.86} & \textbf{76.87} & \textbf{74.68} & \textbf{80.70} & 71.80 & \textbf{66.38} & \textbf{60.18} & \textbf{58.07} & \textbf{65.75} & \textbf{73.60} \\
\bottomrule
\end{tabular}
}
\caption{Accuracy (\%) for each language on Global-MMLU. Within each of the two model groups, the highest value in each column is bold, and the second highest is underlined.}
\label{tab:global_mmlu_results}
\end{table*}

\begin{table*}[t]
\centering
\small
\setlength{\tabcolsep}{3.2pt}
\resizebox{\textwidth}{!}{
\begin{tabular}{l*{13}{c}}
\toprule
& \multicolumn{5}{c}{High} & \multicolumn{3}{c}{Medium} & \multicolumn{4}{c}{Low} & \\
\cmidrule(lr){2-6}\cmidrule(lr){7-9}\cmidrule(lr){10-13}
Model & en & es & fr & zh & ja & ko & id & th & sw & wo & yo & zu & Avg. \\
\midrule
Qwen3-8B & 48.61 & 45.71 & 42.50 & 41.92 & 40.32 & 33.53 & 48.18 & 36.24 & 21.68 & 19.35 & 16.45 & 17.75 & 34.35 \\
Apollo2-7B & 47.89 & 37.41 & 37.99 & 38.86 & 32.17 & 32.89 & 31.44 & 21.39 & 13.10 & 12.22 & 11.93 & 9.75 & 27.25 \\
BioMistral-7B & 25.90 & 22.12 & 22.27 & 13.39 & 9.02 & 8.15 & 15.43 & 7.13 & 13.10 & 14.41 & 12.22 & 13.53 & 14.72 \\
CURE-Med-7B & 40.17 & 31.44 & 32.02 & 37.55 & 27.94 & 26.49 & 31.44 & 23.58 & 14.56 & 16.30 & 14.84 & 14.12 & 25.87 \\
HuatuoGPT-3-8B & 56.33 & 48.18 & 48.18 & 42.06 & 39.30 & 33.92 & 45.12 & 34.49 & 20.81 & 18.05 & 18.77 & 19.94 & 35.43 \\
\rowcolor{LightGreen}
\textbf{\model-8B} & \underline{56.91} & \underline{49.40} & \underline{51.14} & \underline{50.43} & \underline{41.68} & \underline{36.79} & \underline{49.54} & \underline{37.91} & \underline{25.90} & \underline{24.77} & \underline{19.08} & \underline{20.29} & \underline{38.65} \\
\rowcolor{LightGreen}
\textbf{\model-8B w/ \prm} & \textbf{62.73} & \textbf{55.17} & \textbf{56.62} & \textbf{56.91} & \textbf{50.36} & \textbf{47.16} & \textbf{55.17} & \textbf{44.98} & \textbf{28.67} & \textbf{29.69} & \textbf{23.87} & \textbf{26.36} & \textbf{44.81} \\
\midrule
Qwen3.5-9B & 65.50 & 61.42 & 60.69 & 57.81 & 54.87 & 51.55 & 59.24 & 38.28 & 35.95 & 30.13 & 32.60 & 32.31 & 48.36 \\
MedGemma-27B & 60.11 & 58.81 & 59.53 & 52.55 & 49.92 & 44.83 & 55.74 & \underline{51.82} & \underline{45.85} & 13.82 & 28.01 & 35.66 & 46.39 \\
CURE-Med-32B & 67.97 & 62.59 & 62.00 & 58.95 & 52.83 & 50.51 & 58.80 & 47.30 & 29.11 & 26.64 & 19.06 & 21.83 & 46.47 \\
HuatuoGPT-3-32B & 67.54 & 64.34 & 62.29 & 59.09 & 57.49 & 52.83 & 60.69 & \textbf{53.56} & 30.85 & 21.97 & 19.51 & 21.39 & 47.63 \\
Baichuan-M2-32B & 67.68 & 62.73 & 63.60 & 60.69 & 56.91 & 50.50 & 60.69 & 51.23 & 23.87 & 23.14 & 21.54 & 19.21 & 46.82 \\
\rowcolor{LightGreen}
\textbf{\model-9B} & \underline{69.46} & \underline{65.03} & \underline{64.29} & \underline{62.86} & \underline{57.87} & \underline{55.67} & \underline{63.47} & 45.19 & 39.95 & \underline{32.67} & \underline{33.62} & \underline{38.57} & \underline{52.39} \\
\rowcolor{LightGreen}
\textbf{\model-9B w/ \prm} & \textbf{70.31} & \textbf{67.39} & \textbf{67.69} & \textbf{65.07} & \textbf{61.42} & \textbf{59.10} & \textbf{66.23} & 50.07 & \textbf{47.74} & \textbf{38.57} & \textbf{39.88} & \textbf{43.81} & \textbf{56.44} \\
\bottomrule
\end{tabular}
}
\caption{Accuracy (\%) for each language on MMLU-ProX. Within each of the two model groups, the highest value in each column is bold, and the second highest is underlined.}
\label{tab:mmluprox_results}
\end{table*}

\section{Detailed CUREMED-BENCH Results}
\label{sec:appendix_curemed_details}

Table~\ref{tab:curemed_bench_detailed_results} reports the separate CUREMED-BENCH metrics: Logical Accuracy (LA) evaluates the correctness of the final medical answer regardless of its language, while Language Consistency (LC) evaluates whether the final answer uses the question language.

\begin{table*}[t]
\centering
\small
\setlength{\tabcolsep}{2.8pt}
\resizebox{\textwidth}{!}{
\begin{tabular}{l*{8}{c}@{\hspace{0.8em}}*{7}{c}}
\toprule
& \multicolumn{8}{c}{In-Domain} & \multicolumn{7}{c}{Out-of-Domain} \\
\cmidrule(lr){2-9}\cmidrule(lr){10-16}
Model & es & fr & ja & ko & th & sw & yo & Avg. & vi & hi & tr & am & bn & ha & Avg. \\
\midrule
\multicolumn{16}{c}{\textbf{Logical Accuracy}} \\
\midrule
CURE-Med-7B & \underline{80.95} & 72.53 & 67.35 & 50.27 & 62.14 & 42.83 & 35.74 & 58.83 & 67.71 & 47.16 & 32.61 & 23.96 & 51.48 & \underline{49.94} & 45.48 \\
Qwen3-8B & 72.98 & 74.15 & 73.79 & 64.43 & 59.79 & 25.65 & 5.30 & 53.73 & 66.05 & \underline{67.97} & 61.16 & 26.14 & \underline{56.60} & 6.87 & 47.47 \\
HuatuoGPT-3-8B & 35.37 & 34.27 & 31.50 & 28.17 & 31.03 & 14.56 & 5.60 & 25.79 & 34.65 & 36.38 & 20.80 & 15.91 & 30.97 & 5.24 & 23.99 \\
\rowcolor{LightGreen}
\model-8B & 67.33 & 71.47 & 64.81 & 50.43 & 53.45 & 22.30 & 15.11 & 49.27 & 61.89 & 37.73 & 41.19 & 5.47 & 27.87 & 3.20 & 29.56 \\
Qwen3.5-9B & 79.92 & \underline{81.32} & \textbf{77.94} & \textbf{75.72} & \underline{63.63} & \textbf{58.95} & \textbf{52.92} & \textbf{70.06} & \textbf{74.57} & \textbf{77.31} & \textbf{69.61} & \textbf{66.55} & \textbf{64.77} & \textbf{55.32} & \textbf{68.02} \\
\rowcolor{LightGreen}
\model-9B & \textbf{82.48} & \textbf{81.36} & \underline{75.58} & \underline{65.88} & \textbf{64.23} & \underline{43.96} & \underline{42.51} & \underline{65.14} & \underline{71.86} & 60.38 & \underline{62.29} & \underline{32.53} & 40.14 & 25.44 & \underline{48.77} \\
\midrule
\multicolumn{16}{c}{\textbf{Language Consistency}} \\
\midrule
CURE-Med-7B & \underline{80.95} & 79.28 & \underline{81.90} & \underline{69.33} & 27.55 & 25.00 & 0.00 & 52.00 & \underline{51.42} & 39.55 & \underline{71.42} & 32.38 & 35.00 & 0.00 & 38.30 \\
Qwen3-8B & 0.00 & 0.00 & 0.00 & 0.00 & 0.00 & 0.00 & 0.00 & 0.00 & 0.00 & 0.00 & 0.00 & 0.00 & 0.00 & 0.00 & 0.00 \\
HuatuoGPT-3-8B & 0.00 & 0.00 & 0.00 & 0.00 & 0.00 & 0.00 & 0.00 & 0.00 & 0.00 & 0.00 & 0.00 & 0.00 & 0.00 & 0.00 & 0.00 \\
\rowcolor{LightGreen}
\model-8B & 26.67 & \textbf{100.00} & 80.95 & \textbf{97.33} & \textbf{92.85} & \underline{88.57} & \textbf{47.13} & \textbf{76.21} & 25.71 & \textbf{48.63} & 69.64 & \textbf{93.33} & \textbf{99.28} & \underline{13.92} & \underline{58.42} \\
Qwen3.5-9B & 0.00 & 0.00 & 0.00 & 0.00 & 0.00 & 0.00 & 0.00 & 0.00 & 0.00 & 0.00 & 0.00 & 0.00 & 0.00 & 0.00 & 0.00 \\
\rowcolor{LightGreen}
\model-9B & \textbf{100.00} & \underline{99.28} & \textbf{83.81} & 45.33 & \underline{29.59} & \textbf{97.85} & \underline{41.39} & \underline{71.04} & \textbf{97.14} & \underline{41.74} & \textbf{91.07} & \underline{61.91} & \underline{96.42} & \textbf{70.86} & \textbf{76.52} \\
\bottomrule
\end{tabular}
}
\caption{Detailed CUREMED-BENCH results (\%) under zero-shot prompting.}
\label{tab:curemed_bench_detailed_results}
\end{table*}

\begin{table*}[t]
\centering
\small
\setlength{\tabcolsep}{3.0pt}
\resizebox{\textwidth}{!}{
\begin{tabular}{l*{13}{c}}
\toprule
& \multicolumn{5}{c}{High-resource} & \multicolumn{3}{c}{Medium-resource} & \multicolumn{4}{c}{Low-resource} & \\
\cmidrule(lr){2-6}\cmidrule(lr){7-9}\cmidrule(lr){10-13}
Training method & en & es & fr & zh & ja & ko & id & th & sw & wo & yo & zu & Avg. \\
\midrule
\multicolumn{14}{l}{\textbf{Qwen3-8B}} \\
Base model & 68.04 & 64.78 & 60.79 & 62.65 & 53.48 & 54.46 & 65.54 & \textbf{66.80} & 37.54 & 34.60 & 25.90 & 30.71 & 52.11 \\
Stage-one SFT & 75.34 & 66.71 & 66.91 & 68.82 & 57.47 & 56.40 & 53.26 & 56.06 & 31.68 & 23.54 & 23.47 & 29.15 & 50.73 \\
Native SFT & 72.82 & 66.82 & 66.71 & 68.49 & 61.42 & 57.32 & 65.75 & 60.87 & 39.35 & 40.24 & 34.86 & 35.87 & 55.88 \\
Native GRPO & 74.68 & 70.23 & 69.10 & 71.09 & 64.65 & 60.46 & 69.69 & 63.60 & 36.74 & 27.78 & 10.89 & 23.36 & 53.52 \\
Frozen-PRM training & 73.58 & \textbf{71.12} & 70.12 & \textbf{71.69} & 65.14 & \textbf{61.72} & 69.89 & 64.00 & 39.67 & 31.66 & 25.37 & 31.30 & 56.27 \\
\rowcolor{LightGreen}
\textbf{Alternating training (ours)} & \textbf{75.68} & 70.24 & \textbf{70.33} & 71.42 & \textbf{65.58} & 61.46 & \textbf{70.39} & 65.13 & \textbf{44.72} & \textbf{44.62} & \textbf{37.45} & \textbf{37.97} & \textbf{59.58} \\
\midrule
\multicolumn{14}{l}{\textbf{Qwen3.5-9B}} \\
Base model & 80.99 & 76.81 & 76.27 & 76.07 & 71.08 & 67.83 & 77.10 & 61.40 & 60.33 & 51.17 & 48.42 & 53.34 & 66.73 \\
Stage-one SFT & 80.57 & 70.93 & 69.08 & 72.79 & 64.65 & 68.75 & 66.22 & 56.64 & 42.29 & 46.82 & 39.75 & 38.15 & 59.72 \\
Native SFT & 77.51 & 74.40 & 73.76 & 73.97 & 68.91 & 67.30 & 74.23 & 63.67 & 56.93 & \textbf{55.46} & 49.35 & 53.61 & 65.76 \\
Native GRPO & 79.33 & 77.40 & 76.27 & 75.28 & 70.56 & 69.90 & 76.30 & 61.86 & 58.27 & 49.36 & 50.09 & 53.00 & 66.47 \\
Frozen-PRM training & 81.86 & 78.13 & 78.14 & 77.50 & 73.21 & 71.57 & 77.26 & 61.14 & 59.45 & 50.46 & 51.76 & 54.95 & 67.95 \\
\rowcolor{LightGreen}
\textbf{Alternating training (ours)} & \textbf{82.67} & \textbf{79.46} & \textbf{79.53} & \textbf{78.65} & \textbf{74.28} & \textbf{72.19} & \textbf{78.93} & \textbf{64.99} & \textbf{61.71} & 53.05 & \textbf{53.10} & \textbf{58.27} & \textbf{69.74} \\
\bottomrule
\end{tabular}
}
\caption{Accuracy (\%) for each training method and language on Global-MMLU. Bold marks the highest value within each backbone block; ties share the same marking.}
\label{tab:training_methods_global_languages}
\end{table*}

\begin{table*}[t]
\centering
\small
\setlength{\tabcolsep}{3.0pt}
\resizebox{\textwidth}{!}{
\begin{tabular}{l*{13}{c}}
\toprule
& \multicolumn{5}{c}{High-resource} & \multicolumn{3}{c}{Medium-resource} & \multicolumn{4}{c}{Low-resource} & \\
\cmidrule(lr){2-6}\cmidrule(lr){7-9}\cmidrule(lr){10-13}
Training method & en & es & fr & zh & ja & ko & id & th & sw & wo & yo & zu & Avg. \\
\midrule
\multicolumn{14}{l}{\textbf{Qwen3-8B}} \\
Base model & 48.61 & 45.71 & 42.50 & 41.92 & 40.32 & 33.53 & 48.18 & 36.24 & 21.68 & 19.35 & 16.45 & 17.75 & 34.35 \\
Stage-one SFT & \textbf{57.73} & 44.91 & 44.89 & 46.03 & 38.05 & 34.11 & 34.59 & 32.69 & 17.94 & 10.36 & 11.11 & 12.37 & 32.07 \\
Native SFT & 54.10 & 45.40 & 45.74 & 47.52 & 39.57 & 34.64 & 43.25 & 34.78 & 20.61 & 21.92 & 16.89 & 18.05 & 35.21 \\
Native GRPO & 56.62 & 47.88 & 47.16 & 49.78 & 40.32 & \textbf{37.26} & 47.30 & 36.53 & 17.75 & 13.10 & 4.30 & 10.77 & 34.06 \\
Frozen-PRM training & 55.98 & \textbf{50.49} & \textbf{52.54} & \textbf{54.14} & 41.17 & 36.25 & 48.03 & 37.55 & 18.94 & 16.88 & 14.64 & 15.74 & 36.86 \\
\rowcolor{LightGreen}
\textbf{Alternating training (ours)} & 56.91 & 49.40 & 51.14 & 50.43 & \textbf{41.68} & 36.79 & \textbf{49.54} & \textbf{37.91} & \textbf{25.90} & \textbf{24.77} & \textbf{19.08} & \textbf{20.29} & \textbf{38.65} \\
\midrule
\multicolumn{14}{l}{\textbf{Qwen3.5-9B}} \\
Base model & 65.50 & 61.42 & 60.69 & 57.81 & 54.87 & 51.55 & 59.24 & 38.28 & 35.95 & 30.13 & 32.60 & 32.31 & 48.36 \\
Stage-one SFT & 62.64 & 54.49 & 55.21 & 54.57 & 51.73 & 53.94 & 50.61 & 40.96 & 29.76 & 26.20 & 25.38 & 27.68 & 44.43 \\
Native SFT & 60.85 & 56.25 & 56.30 & 54.27 & 50.05 & 47.46 & 55.13 & 40.71 & 35.22 & 32.15 & 30.29 & 32.70 & 45.95 \\
Native GRPO & 65.50 & 59.67 & 60.40 & 56.47 & 52.54 & 49.92 & 58.95 & 38.71 & 35.22 & 27.65 & 29.25 & 31.58 & 47.16 \\
Fronzen-PRM training & 67.46 & 64.13 & 62.48 & 56.79 & 52.35 & 53.47 & 59.83 & 38.42 & 39.75 & 32.56 & 30.29 & 31.19 & 49.06 \\
\rowcolor{LightGreen}
\textbf{Alternating training (ours)} & \textbf{69.46} & \textbf{65.03} & \textbf{64.29} & \textbf{62.86} & \textbf{57.87} & \textbf{55.67} & \textbf{63.47} & \textbf{45.19} & \textbf{39.95} & \textbf{32.67} & \textbf{33.62} & \textbf{38.57} & \textbf{52.39} \\
\bottomrule
\end{tabular}
}
\caption{Accuracy (\%) for each training method and language on MMLU-ProX. Avg. is the unweighted mean over all 12 languages. Bold marks the highest value within each backbone block; ties share the same marking.}
\label{tab:training_methods_prox_languages}
\end{table*}

\section{\prm-Assisted Inference Results}
\label{sec:appendix_prm_stages}

Table~\ref{tab:prm_stage_groups} reports the absolute resource-group accuracies underlying Figure~\ref{fig:prm_stage_comparison}. Tables~\ref{tab:prm_stage_8b_languages} and~\ref{tab:prm_stage_9b_languages} provide the corresponding scores for each language. The 8B and 9B settings use Qwen3-8B and Qwen3.5-9B, respectively, with either Native GRPO or \model policies. Self-consistency selects the majority-vote answer, while Stage-One and Stage-Two PRM rows use best-of-$N$ selection with the corresponding process scorer. All results measure final-answer accuracy on Global-MMLU. 

\begin{table}[t]
\centering
\small
\setlength{\tabcolsep}{3.5pt}
\resizebox{\linewidth}{!}{
\begin{tabular}{l*{4}{c}}
\toprule
Inference strategy & High & Medium & Low & Avg. \\
\midrule
\multicolumn{5}{l}{\textbf{Native GRPO-8B}} \\
Policy alone & 69.95 & 64.58 & 24.69 & 53.52 \\
Self-consistency & 73.54 & 68.33 & 38.45 & 60.54 \\
Stage-One PRM & \textbf{75.55} & 71.20 & 39.48 & 62.44 \\
Stage-Two PRM & 75.27 & \textbf{71.49} & \textbf{41.08} & \textbf{62.93} \\
\midrule
\multicolumn{5}{l}{\textbf{\model-8B}} \\
Policy alone & 70.65 & 65.66 & 41.19 & 59.58 \\
Self-consistency & 74.51 & 70.11 & 44.91 & 63.54 \\
Stage-One PRM & 75.40 & 70.65 & 42.85 & 63.36 \\
Stage-Two PRM & \textbf{76.41} & \textbf{71.71} & \textbf{45.66} & \textbf{64.99} \\
\midrule
\multicolumn{5}{l}{\textbf{Native GRPO-9B}} \\
Policy alone & 75.77 & 69.35 & 52.68 & 66.47 \\
Self-consistency & 78.31 & 72.53 & 58.08 & 70.12 \\
Stage-One PRM & \textbf{79.86} & \textbf{75.04} & 60.26 & 72.12 \\
Stage-Two PRM & 79.76 & 74.95 & \textbf{60.98} & \textbf{72.30} \\
\midrule
\multicolumn{5}{l}{\textbf{\model-9B}} \\
Policy & 78.92 & 72.04 & 56.53 & 69.74 \\
Self-consistency & 80.09 & 74.02 & 59.20 & 71.61 \\
Stage-One PRM & 80.35 & 74.22 & 59.04 & 71.71 \\
Stage-Two PRM & \textbf{81.13} & \textbf{75.73} & \textbf{62.60} & \textbf{73.60} \\
\bottomrule
\end{tabular}
}
\caption{Different inference accuracy (\%) on Global-MMLU by resource group. Bold marks the highest value among inference strategies for the same policy.}
\label{tab:prm_stage_groups}
\end{table}

\begin{table}[t]
\centering
\small
\setlength{\tabcolsep}{2pt}
\begin{tabular*}{\linewidth}{@{\extracolsep{\fill}}l*{6}{r}@{}}
\toprule
& \multicolumn{3}{c}{\model-8B} & \multicolumn{3}{c}{\model-9B} \\
\cmidrule(lr){2-4}\cmidrule(lr){5-7}
Language & GPT & PRM-8B & $\rho$ & GPT & PRM-9B & $\rho$ \\
\midrule
en & 0.855 & 0.742 & 0.244 & 0.871 & 0.736 & 0.462 \\
es & 0.727 & 0.695 & 0.433 & 0.795 & 0.735 & 0.411 \\
fr & 0.707 & 0.668 & 0.511 & 0.777 & 0.684 & 0.485 \\
zh & 0.780 & 0.718 & 0.412 & 0.824 & 0.716 & 0.340 \\
ja & 0.669 & 0.703 & 0.334 & 0.738 & 0.723 & 0.315 \\
\midrule
ko & 0.600 & 0.615 & 0.391 & 0.750 & 0.657 & 0.514 \\
id & 0.721 & 0.691 & 0.325 & 0.839 & 0.715 & 0.527 \\
th & 0.635 & 0.679 & 0.351 & 0.603 & 0.629 & 0.478 \\
\midrule
sw & 0.347 & 0.645 & 0.241 & 0.561 & 0.646 & 0.479 \\
wo & 0.488 & 0.670 & 0.284 & 0.498 & 0.588 & 0.441 \\
yo & 0.325 & 0.668 & 0.088 & 0.429 & 0.594 & 0.204 \\
zu & 0.253 & 0.635 & 0.237 & 0.517 & 0.638 & 0.326 \\
\bottomrule
\end{tabular*}
\caption{Reasoning quality and \prm agreement for each language.}
\label{tab:reasoning_quality_languages}
\end{table}

\begin{table*}[t]
\centering
\small
\setlength{\tabcolsep}{3.0pt}
\resizebox{\textwidth}{!}{
\begin{tabular}{l*{13}{c}}
\toprule
& \multicolumn{5}{c}{High-resource} & \multicolumn{3}{c}{Medium-} & \multicolumn{4}{c}{Low} & \\
\cmidrule(lr){2-6}\cmidrule(lr){7-9}\cmidrule(lr){10-13}
Inference strategy & en & es & fr & zh & ja & ko & id & th & sw & wo & yo & zu & Avg. \\
\midrule
\multicolumn{14}{l}{\textbf{Native GRPO-8B}} \\
Policy alone & 74.68 & 70.23 & 69.10 & 71.09 & 64.65 & 60.46 & 69.69 & 63.60 & 36.74 & 27.78 & 10.89 & 23.36 & 53.52 \\
Self-consistency & 78.47 & 73.55 & 73.22 & 73.88 & 68.57 & 63.98 & 73.09 & 67.93 & 42.52 & 38.41 & 34.08 & 38.79 & 60.54 \\
Stage-One PRM & \textbf{79.66} & \textbf{74.42} & \textbf{75.38} & \textbf{77.60} & \textbf{70.69} & 67.71 & 75.70 & 70.20 & 45.58 & 39.61 & 33.55 & 39.18 & 62.44 \\
Stage-Two PRM & 79.26 & \textbf{74.42} & 74.55 & 77.47 & 70.63 & \textbf{68.11} & \textbf{76.03} & \textbf{70.33} & \textbf{46.77} & \textbf{42.35} & \textbf{35.02} & \textbf{40.18} & \textbf{62.93} \\
\midrule
\multicolumn{14}{l}{\textbf{\model-8B}} \\
Policy & 75.68 & 70.24 & 70.33 & 71.42 & 65.58 & 61.46 & 70.39 & 65.13 & 44.72 & 44.62 & 37.45 & 37.97 & 59.58 \\
Self-consistency & 79.00 & 73.82 & 73.82 & 75.28 & 70.63 & 67.11 & 73.89 & 69.33 & \textbf{48.63} & \textbf{48.76} & 40.00 & 42.25 & 63.54 \\
Stage-One PRM & 79.93 & 75.54 & 74.41 & 76.48 & 70.63 & 67.74 & 74.69 & \textbf{69.53} & 44.85 & 46.16 & 39.33 & 41.04 & 63.36 \\
Stage-Two PRM & \textbf{80.33} & \textbf{75.81} & \textbf{75.88} & \textbf{77.54} & \textbf{72.49} & \textbf{69.17} & \textbf{76.43} & \textbf{69.53} & 48.30 & 47.96 & \textbf{42.25} & \textbf{44.13} & \textbf{64.99} \\
\bottomrule
\end{tabular}
}
\caption{Different inference accuracy (\%) for each language on Global-MMLU using 8B policies.}
\label{tab:prm_stage_8b_languages}
\end{table*}

\begin{table*}[t]
\centering
\small
\setlength{\tabcolsep}{3.0pt}
\resizebox{\textwidth}{!}{
\begin{tabular}{l*{13}{c}}
\toprule
& \multicolumn{5}{c}{High} & \multicolumn{3}{c}{Medium} & \multicolumn{4}{c}{Low} & \\
\cmidrule(lr){2-6}\cmidrule(lr){7-9}\cmidrule(lr){10-13}
Inference strategy & en & es & fr & zh & ja & ko & id & th & sw & wo & yo & zu & Avg. \\
\midrule
\multicolumn{14}{l}{\textbf{Native GRPO-9B}} \\
Policy alone & 79.33 & 77.40 & 76.27 & 75.28 & 70.56 & 69.90 & 76.30 & 61.86 & 58.27 & 49.36 & 50.09 & 53.00 & 66.47 \\
Self-consistency & 82.25 & 78.73 & 78.67 & 78.00 & 73.88 & 71.82 & 78.70 & 67.06 & 62.65 & 55.24 & 55.68 & 58.74 & 70.12 \\
Stage-One PRM & \textbf{83.12} & 79.73 & \textbf{80.73} & \textbf{79.47} & 76.27 & 74.48 & 79.71 & \textbf{70.93} & 65.78 & 57.98 & \textbf{57.07} & 60.21 & 72.12 \\
Stage-Two PRM & 82.45 & \textbf{80.53} & 80.53 & 78.87 & \textbf{76.41} & \textbf{74.61} & \textbf{79.77} & 70.47 & \textbf{66.51} & \textbf{59.58} & 55.61 & \textbf{62.21} & \textbf{72.30} \\
\midrule
\multicolumn{14}{l}{\textbf{\model-9B}} \\
Policy alone & 82.67 & 79.46 & 79.53 & 78.65 & 74.28 & 72.19 & 78.93 & 64.99 & 61.71 & 53.05 & 53.10 & 58.27 & 69.74 \\
Self-consistency & \textbf{84.38} & 79.86 & 80.93 & 79.47 & 75.81 & 74.02 & 79.77 & 68.26 & 63.12 & 56.51 & 55.74 & 61.41 & 71.61 \\
Stage-One PRM & 83.98 & 80.59 & 80.73 & 79.86 & 76.61 & 73.42 & 80.10 & 69.13 & 64.05 & 56.05 & 54.16 & 61.88 & 71.71 \\
Stage-Two PRM & 84.32 & \textbf{81.59} & \textbf{81.99} & \textbf{80.86} & \textbf{76.87} & \textbf{74.68} & \textbf{80.70} & \textbf{71.80} & \textbf{66.38} & \textbf{60.18} & \textbf{58.07} & \textbf{65.75} & \textbf{73.60} \\
\bottomrule
\end{tabular}
}
\caption{Differnt inference accuracy (\%) for each language on Global-MMLU using 9B policies.}
\label{tab:prm_stage_9b_languages}
\end{table*}

\section{Training Method Results by Language}
\label{sec:appendix_training_methods}

Tables~\ref{tab:training_methods_global_languages} and~\ref{tab:training_methods_prox_languages} report accuracy by language.  
In particular, frozen-PRM training uses the frozen Stage-One PRM for three policy-training rounds.

\paragraph{Language-specific variation.}
Alternating training achieves the highest accuracy in 7 of 12 languages at 8B and all 12 at 9B on MMLU-ProX; on Global-MMLU, it leads in 8 of 12 languages at 8B and 11 of 12 at 9B. 
Relative to the original backbones, it improves accuracy in 47 of the 48 benchmark--backbone--language combinations. 
The remaining combination is Thai on Global-MMLU at 8B, where accuracy changes from 66.80\% to 65.13\%. 
These individual-language results complement the improvements over the base models in all three resource-group averages.

\section{Reasoning and PRM Quality }
\label{sec:appendix_reasoning_quality}

For each of \model-8B and \model-9B, we sample 50 outputs in each language from Global-MMLU and 50 from MMLU-ProX, giving 100 outputs for each language and 2,400 across both policies. 
The \texttt{GPT-6.1-sol} judge evaluates each reasoning step for medical correctness and clinical or scientific logic on $[0,1]$. 
A score of $0$ denotes a clear error; $0.5$ denotes partial correctness, ambiguity, or insufficient context; and $1$ denotes a medically correct and logically sound step. 
Intermediate values are allowed. 
Each output is also scored by the corresponding Stage-Two \prm.

\paragraph{Rank agreement.}
For each answer, we separately average its LLM-judge step scores and its Stage-Two \prm step scores. Within each model--language--benchmark setting, we compute Spearman correlation across the 50 pairs of answer-level means. 
The language-level $\rho$ is the equal-weight average of the Global-MMLU and MMLU-ProX correlations. 
Table~\ref{tab:reasoning_quality_languages} reports language-level mean scores and these benchmark-averaged correlations. 

\begin{tcolorbox}[
  breakable,
  width=\linewidth,
  title={LLM-judge prompt for reasoning quality},
  fonttitle=\small\bfseries,
  fontupper=\small,
  coltitle=black,
  colbacktitle=black!8,
  colback=black!2,
  colframe=black!35,
  boxrule=0.5pt,
  arc=1mm,
  left=6pt,
  right=6pt,
  top=5pt,
  bottom=5pt
]
You are a medical reasoning evaluator. Analyze the supplied reasoning steps and assess the medical correctness and clinical or scientific logic of each step.

Assign one validity score per step, in the original order:

- $0$: The step contains a clear medical, scientific, or logical error.\par

- $0.5$: The step is partially correct, ambiguous, or cannot be established from the available context.\par

- $1$: The step is medically correct and logically sound.

You may use any numeric value between $0$ and $1$ to express intermediate validity.
\end{tcolorbox}

\section{Language Consistency Reward}
\label{sec:appendix_language_consistency}

\paragraph{Classifier Construction.}
We fit a character n-gram classifier on the training prompts and their language labels. 
The classifier uses character 3-, 4-, and 5-grams to compute an average log probability from smoothed frequencies for each language. 
A softmax with temperature $0.05$ converts these scores into language probabilities $P(\ell\mid x)$ for a text chunk $x$.

\paragraph{Reasoning-Text Preprocessing.}
For rollout scoring, we use only the parsed \texttt{<step>} contents. 
We normalize this text by retaining Unicode letters, combining marks, and whitespace, then divide the normalized text into 240-character chunks. 
Chunks containing fewer than 20 letters are discarded.

\paragraph{Rollout Scoring.}
For the requested language $\ell$, we average its predicted probability over qualifying chunks, weighted by their letter counts:
\begin{equation}
    R_{\mathrm{lang}}(\tau)
    = \frac{\sum_{x\in\mathcal{C}(\tau)}n_x\,P(\ell\mid x)}
           {\sum_{x\in\mathcal{C}(\tau)}n_x},
    \label{eq:language_reward}
\end{equation}
where $\mathcal{C}(\tau)$ is the set of qualifying chunks and $n_x$ is the number of letters in chunk $x$. 
We set this reward to zero when no qualifying chunk remains.

\section{Semantic Alignment Reward}
\label{sec:appendix_semantic_alignment}

\paragraph{English Teacher Selection.}
We select a teacher from the current multilingual rollout group for the same underlying question. 
Eligible teachers are English rollouts with a correct final answer, a valid output format, and at least one parsed reasoning step. 
Among these candidates, we choose the rollout with the highest mean \prm score.

\paragraph{Step Encoding.}
We encode each reasoning step with the frozen \texttt{ibm-granite/}\allowbreak\texttt{granite-embedding-311m-}\allowbreak\texttt{multilingual-r2} model. 
Each step is truncated to at most 512 tokens. We retain the first 256 dimensions of its CLS embedding and apply L2 normalization.

\paragraph{Alignment Scoring.}
For non-empty target and teacher reasoning sequences with $m$ and $n$ steps, respectively, let $C_{tu}$ be the cosine similarity between target step $t$ and teacher step $u$. 
The directional means and rescaling used in Equation~\eqref{eq:alignment_reward} are
\begin{equation}
\begin{aligned}
    S_{\mathrm{tgt}}
    &= \frac{1}{m}\sum_{t=1}^{m}\max_{1\leq u\leq n} C_{tu}, \\
    S_{\mathrm{en}}
    &= \frac{1}{n}\sum_{u=1}^{n}\max_{1\leq t\leq m} C_{tu}, \\
    \operatorname{scale}(s)
    &= \operatorname{clip}\!\left(\frac{s-0.55}{0.80-0.55},0,1\right).
\end{aligned}
    \label{eq:alignment_details}
\end{equation}
This comparison accommodates different step counts and imposes no ordering constraint. 
The language-related product term is zero for English rollouts or when no eligible English teacher is available. 
A target-language rollout can receive this bonus even when its final answer is incorrect.

\end{document}